\relax
\documentclass[letterpaper]{article} 
\usepackage{aaai21}  
\usepackage{bm}
\usepackage[T1]{fontenc}
\usepackage{amsmath}
\usepackage{amsfonts}
\usepackage{mathtools}
\usepackage{times}  
\usepackage{helvet} 
\usepackage{courier}  
\usepackage[hyphens]{url}  
\usepackage{algorithm}
\usepackage{algorithmicx}
\usepackage{algpseudocode}
\usepackage{graphicx}
\usepackage{natbib}  
\usepackage{caption} 
\DeclareMathOperator*{\argmax}{arg\,max}
\DeclareMathOperator*{\argmin}{arg\,min}

\title{Reinforcement Learning In Two Player Zero Sum Simultaneous Action Games}
\author{
  Patrick Phillips
  
}
\affiliations{
    250 Hutchison Rd, Rochester, NY 14620
    
    Correspondence to pphill10@u.rochester.edu
}

\begin{document}
\maketitle

\begin{abstract}
Two player zero sum simultaneous action games are common in video games, financial markets, war, business competition, and many other settings.
We first introduce the fundamental concepts of reinforcement learning in two player zero sum simultaneous action games and discuss the unique challenges this type of game poses. Then we introduce two novel agents that attempt to handle these challenges by using joint action Deep Q-Networks (DQN).
The first agent, called the Best Response AgenT (BRAT), builds an explicit model of its opponent's policy using imitation learning, and then uses this model to find the best response to exploit the opponent's strategy. 
The second agent, Meta-Nash DQN, builds an implicit model of its opponent's policy in order to produce a context variable that is used as part of the Q-value calculation. An explicit minimax over Q-values is used to find actions close to Nash equilibrium.
We find empirically that both agents converge to Nash equilibrium in a self-play setting for simple matrix games, while also performing well in games with larger state and action spaces. 
These novel algorithms are evaluated against vanilla RL algorithms as well as recent state of the art multi-agent and two agent algorithms.
This work combines ideas from traditional reinforcement learning, game theory, and meta learning. 
\end{abstract}

\section{Introduction}
\noindent An important class of games to consider in reinforcement learning is two player simultaneous action games. Common examples include Rock-Paper-Scissors, Pong, predator-prey games, iterated matrix games, and many real life scenarios such as business competition or war. Furthermore, as we show later, many team games in which agents on two teams share the same objective can also be reformulated into this setting. These types of games include sports such as soccer and basketball as well as video games such as DOTA or Overwatch.

Two player simultaneous action games, and more generally multi-agent games, pose some unique challenges. The most immediate and fundamental challenge is that from the view of any particular agent, the environment is non-stationary. This is known as the moving-target problem ~\cite{hernandez-leal_survey_2019, al-shedivat_continuous_2018, bansal_emergent_2018}. Another difficulty is that of credit assignment; how can any particular agent know if it was their good actions or just their opponents mistakes that lead to the reward they received? A more universal challenge in reinforcement learning and machine learning in general is the curse of dimensionality. This difficulty is exacerbated in multi-agent games since additional agents create exponentially larger state and action spaces~\cite{andriotis2018managing}. One final challenge is that of policy robustness. It is difficult to ensure that any particular action or policy will work well regardless of the other agent's behavior~\cite{hernandez-leal_survey_2019}. In some games it is quite often the case that a particular action will only be good against a particular opponent. Agents can stick to playing a Nash Equilibrium strategy, however this can often lead to overcautious behavior that has little chance of doing better than breaking even. Particularly in two player games, where much more effort can be put into modelling one's opponent, finding and acting on opponent's weaknesses is often viable and effective. Learning robust policies, while also exploiting the weaknesses of opponents is a key motivation for the algorithms we present.

\section{MDP Formalization}
A single agent Markov Decision Process (MDP) is defined by the tuple $\langle \mathcal{S}, \mathcal{A}, \mathcal{T}, \mathcal{R}, \gamma \rangle$, where $\mathcal{S}$ represents some finite set of states and $\mathcal{A}$ represents a finite set of actions. The transition function $\mathcal{T} : \mathcal{S} \times \mathcal{A} \times \mathcal{S} \rightarrow [0,1]$ gives the probability of the transition from state $s\in \mathcal{S}$ on action $a \in \mathcal{A}$ to next state $s' \in \mathcal{S}$. The reward function $\mathcal{R} : \mathcal{S} \times \mathcal{A} \times \mathcal{S} \rightarrow \mathrm{R} $ gives the (possibly stochastic) reward that an agent receives from being in state $s$ taking action $a$ and ending in state $s'$. Finally $\gamma \in [0,1]$ is a discount factor that indicates how much less important future rewards are than immediate rewards.

MDPs can be simply extended to multi-agent MDPs that are defined as the tuple $\langle \mathcal{S}, \mathcal{N}, \mathcal{A}, \mathcal{T}, \mathcal{R}\rangle$. The key distinctions are that now the action space, rewards, and transition function are given for the joint set of agents, formally we write this as $\mathcal{A} = \mathcal{A}_1 \times \mathcal{A}_2 \times ... \times \mathcal{A}_N$,  $\mathcal{R} = \mathcal{R}_1 \times \mathcal{R}_2 \times ... \times \mathcal{R}_N$, and $\mathcal{T} = \mathcal{S} \times \mathcal{A}_1 \times \mathcal{A}_2 \times ... \times \mathcal{A}_N$. 

We consider a particular class of multi-agent MDPs called two player zero sum games where $\mathcal{N}=2$ and $\mathcal{R}_1=-\mathcal{R}_2$. Note that certain games with more agents can be reformulated as a two player zero sum MDP if there are always only two distinct rewards $\mathcal{R}_1$ and $\mathcal{R}_2$ received by all the agents and $\mathcal{R}_1=-\mathcal{R}_2$. This includes games such as multi-agent soccer, DOTA, and more. The reformulation can be achieved by setting the action space for the first agent to be the joint action $\mathcal{A}_1' = \mathcal{A}_1 \times \mathcal{A}_2 \times ... \times \mathcal{A}_M$ for all agents who receive rewards $\mathcal{R}_1$ and similarly setting the joint action $\mathcal{A}'_2  = \mathcal{A}_1 \times \mathcal{A}_2 \times ... \times \mathcal{A}_P$ for agents who receive the rewards $\mathcal{R}_2$. Sometimes handling this class of multi-agent MDPs with joint actions can be beneficial, as it helps to better coordinate a team's actions, however sometimes the joint action space of size $|A|^\mathcal{M}$ is far too large to consider as it grows exponentially with $M$, the number of agents on a team.

\section{Relevant Game Theory}
The first key notion from game theory that we rely heavily on is that of Nash Equilibrium. A set of policies $\boldsymbol\pi(s) = \{\pi_i(s) \}_{i\in N}$ form a Nash Equilibrium if unilateral deviation from this equilibrium by a single agent cannot improve the value of that agent's policy. The value of a policy we view as the sum of discounted rewards, which we define for each agent $i$ as

\begin{equation}
    \mathcal{R}_i\left( s ; \pi_i , \boldsymbol\pi^*_{-i} \right) = \mathrm{E}\left[\sum_{t=0}^\infty \gamma^t r_i(s_t,\pi_{i}(s_t),\boldsymbol\pi_{-i}(s_t)\right]
\end{equation}

Now formally, a collection of policies $\boldsymbol\pi(s)$ forms a Nash equilibrium if 
\begin{equation} 
\label{eq:Nash-equilibrium-def}
	\mathcal{R}_i\left( s ; \pi_i , \boldsymbol\pi^*_{-i} \right)
	\leq
	\mathcal{R}_i\left( s ; \pi^*_i , \boldsymbol\pi^*_{-i} \right)
\end{equation}
for all states $s$ and admissible policies $\pi_i$ and for all $i\in N$. In Eq. \ref{eq:Nash-equilibrium-def} we use $\pi^*$ to denote that the policies are that of a Nash Equilibrium, and the notation $\pi_{-i}$ to denote the set of policies for all agents excluding the $i$-th agent. 

One particular type of game that is of significant importance to subsequent sections are matrix games, and more specifically zero sum matrix games. A matrix game is defined uniquely by a payoff matrix $\mathbf{A}$ such as 
\begin{table}[h!]
\hspace{.2cm} $\mathbf{A}$ = \hspace{.5cm}
\begin{tabular}{cc|c|c|c|}
  & \multicolumn{1}{c}{} & \multicolumn{3}{c}{Player $2$} \\
  & \multicolumn{1}{c}{} & \multicolumn{1}{c}{$A$}  & \multicolumn{1}{c}{$B$}  & \multicolumn{1}{c}{$C$} \\\cline{3-5}
            & $A$ & $(x,y)$ & $(x,y)$ & $(x,y)$ \\ \cline{3-5}
Player $1$  & $B$ & $(x,y)$ & $(x,y)$ & $(x,y)$ \\\cline{3-5}
            & $C$ & $(x,y)$ & $(x,y)$ & $(x,y)$ \\\cline{3-5}
\end{tabular}
\end{table}
\newline
In the matrix game Player 1 chooses the a row, and Player 2 chooses a column. They then receive rewards according to the entry corresponding to the choice of row and column, where Player 1 receives reward $x$ and Player 2 receives $y$. The game is zero sum if $x=-y$ for all entries, and in this case the game can be specified by only a single entry at each location of the matrix. Each player is generally allowed to use a mixed strategy which is a probability distribution over available actions. The extension of matrix games to $n$-agents is a straightforward process which uses payoffs given by $n$-dimensional tensors.

John Nash proved that any game with a finite number of players each allowed to use a mixed strategy over a finite set of actions has at least one Nash Equilibrium~\cite{nash_equilibrium_1950}. However, it has since been proved that finding if there exists a second equilibrium point is NP-Complete~\cite{daskalakis2009complexity}.  There are many algorithms to find the Nash Equilibria of matrix games, however they all take exponential time in the worst case. One particularly useful algorithm for our purposes however is the Lemke-Howson algorithm which efficiently finds \textit{one} equilibrium point (it also takes worst case exponential time, but practically runs much faster than this)~\cite{lemke_equilibrium_nodate}. We use the Lemke-Howson algorithm to compute Nash Equilibrium of matrix games throughout this work. For a simple example of Nash Equilibrium, consider the Prisoner's Dilemma where each player can either cooperate (\textit{Co}) or defect (\textit{Def}). 
\begin{table}[h!]
\hspace{.5cm}
\begin{tabular}{cc|c|c|c|}
  & \multicolumn{1}{c}{} & \multicolumn{3}{c}{Player $2$} \\
  & \multicolumn{1}{c}{} & \multicolumn{1}{c}{$Co$}  & \multicolumn{1}{c}{$Def$} \\\cline{3-4}
            & $Co$ & $(-1,-1)$ & $(-3,0)$ \\\cline{3-4}
Player $1$  & $Def$ & $(0,-3)$ & $(-2,-2)$ \\\cline{3-4}
\end{tabular}
\end{table}
\newline
Consider the two strategies in which each player plays \textit{Def}; if either agent were to unilaterally deviate from this strategy they would be worse off. Thus this is the Nash Equilibrium, despite the counter-intuitive fact that both parties would be better off if they cooperated. 

Iterated matrix games, which are used as the environment for some of our experiments, consist of repeatedly playing a matrix game. The state in these iterated matrix games is encoded as the most recent set of actions. In multi-stage games such as iterated matrix games, there are many more complex notions of equilibrium points that go beyond Nash Equilibrium since unilateral deviation is an unreasonable basis. The most fundamental equilibrium point, called the Subgame perfect Nash Equilibrium, describes a set of strategies in multi-stage games that are Nash Equilibrium strategies in every subgame, where a subgame is any smaller part of the multi-stage game~\cite{harsanyi1988general}. For games with only a single Nash equilibrium point such as the Prisoner's Dilemma, the Subgame perfect Nash Equilibrium is again to always defect, even though this makes both players much worse off than cooperating, especially for games with many iterations. However, since the games we consider are zero-sum, we do not have to contend with handling counter-intuitive equilibria like this one.

\section{Deep Q-Network (DQN)}
The explosion of reinforcement learning has been largely inspired by the 2015 paper by Mnih et. al. that introduced Deep Q-Networks (DQN)~\cite{mnih_human-level_2015}. DQN learns to simply minimize the loss function
\begin{equation}
\label{eq:TD-loss}
    L_i(\theta_i) = \mathrm{E}_{(s,a,s',r) \sim D} \left[\left(Q(s,a;\theta_i) - r + \gamma\max_{a'}Q(s',a',\theta^-_i)\right)\right]
\end{equation}
where $\theta_i$ denotes the parameters of the Q-Network at iteration $i$, and $\theta^-_i$ denotes the parameters of a target Q-network that is periodically updated to the current Q-Network. This loss function, sometimes called the TD(0) loss, is a one step bootstrap that was first inspired by Bellman's steady state equation for the Q-function. The term
\begin{equation}
\label{eq:td-target}
    r + \gamma\max_{a'}Q(s',a',\theta^-_i)
\end{equation} 
is referred to as the target or TD-target, since this is the target that we are chasing with our updates to parameters $\theta$. During optimization of the loss function, minibatches of experience which consist of transitions $(s,a,s',r)$ are drawn from the replay buffer $D$. The use of this replay buffer from which random samples are drawn stands in contrast to prior work in which updates to the Q function were done online (in sequential order of experience). Using a replay buffer to decorrelate transitions, and also using a target network to compute targets to avoid a non-stationarity problem were two pivotal contributions of Mnih et. al.'s DQN.

\section{Minimax-DQN}
A Minimax Q-Learning algorithm was proposed in 1994 by Michael Littman~\cite{littman_markov_1994}. This algorithm explicitly calculates min-max values of matrix games and does exact tabular updates. Recent work has extended Littman's algorithm to make use of function approximation similarly to DQN~\cite{fan_theoretical_2020} in an algorithm called Minimax-DQN. The Minimax-DQN algorithm is the building block of our RL agents.

\begin{figure}
    \centering
    \includegraphics[width=0.45\textwidth]{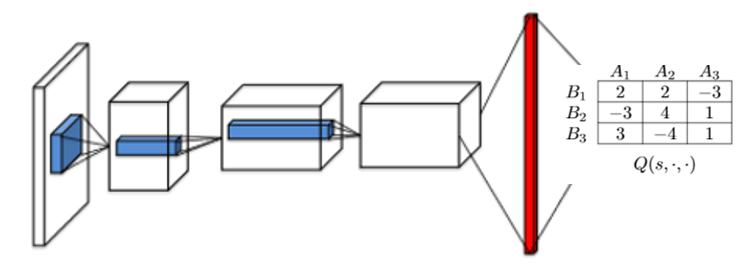}
    \caption{Minimax-DQN architecture}
    \label{fig:Minimax-DQN}
\end{figure}

The key insight that Littman had in his 1994 work was that at any particular state $s \in \mathcal{S}$ the Q-function $Q(s, \cdot, \cdot)$ can be thought of as a matrix game where the entries represent the values of being in state $s$, and where the first agent selects a row action and the second agent a column action. Since the first agent is trying to maximize his rewards, and the second agent is trying to minimize the first agent's rewards (maximize his own rewards), we can take the Nash Equilibrium value of $Q(s, \cdot, \cdot)$ with a min-max calculation, and then use this value to find the TD-target. This yields the intuitive extension of DQN algorithms to simply adapt the TD(0) loss by replacing target values of Eq. \ref{eq:td-target} with
\begin{equation}
    \label{eq:minimax-target}
    y_i = r_t + \gamma \cdot \underset{a}{\max} \underset{b}{\min} \mathop{\mathbb{E}} \left[ Q_{\theta}^{-}(s_{t+1}, a, b) \right].
\end{equation}
While Littman originally proposed this TD-target for tabular values, it is also useful when using function approximators like neural networks. More specifically, a Deep-Q-Network typically takes in the state and outputs the value for taking any given action from that state. Thus to use this minimax-target, we use a Minimax-DQN that outputs a matrix of values as shown in Figure \ref{fig:Minimax-DQN}. The loss is then given analogously to Equation \ref{eq:TD-loss}, by simply substituting in the TD-target given in Equation \ref{eq:td-target}. Algorithm 1 shows the full pseudocode for Minimax-DQN utilizing minimax target values.

\begin{algorithm}[h!]
\begin{algorithmic}
\State \textbf{Input}: A two player zero-sum Markov game ($\mathcal{S}, \mathcal{A}, \mathcal{B}, P, R, \gamma$), replay buffer $\mathcal{D}$, minibatch size $n$, set of opponent policies $\pi_i$, and architecture of minimax deep Q-network $Q_\theta: \mathcal{S} \xrightarrow{}{}\mathcal{A} \times \mathcal{B}. $
\State Initialize replay memory $\mathcal{D}$ to capacity $N$
\State Initialize action-value network $Q_\theta$ with random weights $\theta$
\State Initialize target network $Q_\theta^-$ = $Q_\theta$ 
\While{not done}
    \For {$k=1,K$}
    	\State Set $a_t, b_t= \underset{b_t}{\text{argmax}}\hspace{.1cm}\underset{a_t}{\text{argmin}} Q_\theta(s_t, a_t, b_t)$
    	\State Execute actions $a_t$ and $b_t$ and observe reward $r_t$ and state $s_{t+1}$
    	\State Store transition $\left(s_t, a_t, b_t, r_t, s_{t+1}\right)$ in $\mathcal{D}_i$
    \EndFor
	\State Sample minibatch of transitions 
	\State $\left(s_t, a_t, b_t, r_t, s_{t+1}\right)_{i \in [n]}$ from $\mathcal{D}_i$
	\State Set target
	\State \hspace{1cm} $y_i = r_{t,i} + \gamma \cdot \underset{a}{\max}\hspace{.1cm}\underset{b}{\min}  \left[Q_\theta^-(s_{t+1, i}, a, b)\right]$
	\State using target network $Q^-_\theta$. Then optimize parameters using loss
    \State \hspace{1cm}	$L(\theta) = \frac{1}{n}\sum_{i \in |n|} \left(y_i-Q_\theta(s_{t,i}, a_{t,i}, b_{t,i})\right)^2 $
\EndWhile
\end{algorithmic}
\caption{Minimax-DQN}
\label{alg}
\end{algorithm} 
\section{Best Response AgenT (BRAT)}
The first agent we introduce is the Best Response AgenT BRAT. BRAT uses Minimax-DQN updates to learn a Q-function while simultaneously building a model of its opponent's policy using imitation learning. Ideally a few-shot or one-shot imitation learning algorithm would be used, however for simplicity we implement the imitation learning using a simple classification NN that is trained on (state, action) pairs from its opponent. 

This naive form of imitation learning, often called behavioral cloning, suffers from compounding error caused by covariate shift~\cite{ross2010efficient}. Essentially, once the imitation learner makes one mistake, it will now be working with state spaces that come from outside the distribution of data it trained on, and continue to make more mistakes and progress further from its training distribution. For the small state spaces that we run experiments on, behavioral cloning proves sufficient. 
\begin{algorithm}[h!]
\begin{algorithmic}
\State \textbf{Input}: A two player zero-sum Markov game ($\mathcal{S}, \mathcal{A}, \mathcal{B}, P, R, \gamma$), replay buffer $\mathcal{D}$, minibatch size $n$, opponent policy $\pi^{opp}$, and architecture of minimax deep Q-network $Q_\theta: \mathcal{S} \xrightarrow{}{}\mathcal{A} \times \mathcal{B}. $ 
\State Initialize opponent policy model to $\pi^{opp}_{\phi}$ 
\State Initialize replay memory $\mathcal{D}$ to capacity $N$
\State Initialize action-value network $Q_\theta$ with random weights $\theta$
\State Initialize target network $Q_\theta^-$ = $Q_\theta$ 
\While{not done}
    \For {$k=1,K$}
    	\State Select actions
    	\State \hspace{1cm} $a_t = \underset{a_t}{\argmax}$\hspace{.2cm}$\pi^{opp}_{\phi}(s_t) Q_\theta(s_t, a_t, \cdot)$
    	\State \hspace{1cm} $b_t = \pi^{opp}(s_t)$
    	\State Execute actions $a_t$ and $b_t$ and observe $r_t$ and $s_{t+1}$
    	\State Store transition $\left(s_t, a_t, b_t, r_t, s_{t+1}\right)$ in $\mathcal{D}_i$
    \EndFor
	\State Sample minibatch of transitions
	\State $\left(s_t, a_t, b_t, r_t, s_{t+1}\right)_{i \in [n]}$ from $\mathcal{D}_i$
	\State Set the target
	\State \hspace{.1cm} $y_i = r_{t,i} + \gamma \cdot \underset{a}{\max}  \left[ \pi^{opp}_{\phi}(s_{t+1,i}) Q_\theta^-(s_{t+1, i}, a,\cdot)\right]$
	\State \hspace{3cm} OR
	\State \hspace{.1cm} $y_i = r_{t,i} + \gamma \cdot \underset{a}{\max} \underset{b}{\min}  \left[ Q_\theta^-(s_{t+1, i}, a,b)\right]$
	\State using target network $Q^-_\theta$. Then optimize parameters
	\State with loss:
    \State \hspace{1cm}	$L(\theta) = \frac{1}{n}\sum_{i \in |n|} \left(y_i-Q_\theta(s_{t,i}, a_{t,i}, b_{t,i})\right)^2 $
    \State Update ${\phi}$ using log-loss for (state, action) pairs:
    \State $L(\phi) = -\frac{1}{N} \sum_{i=1}^N \sum_{j=1}^M \mathrm{I}(a_j = a_{t,i}) \log \pi^{opp}_{\phi}(s_{t,i}) $
\EndWhile
\end{algorithmic}
\caption{Best Response AgenT (BRAT)}
\label{alg:BRAT}
\end{algorithm} 

\begin{figure}[h!]
    \centering
    \includegraphics[width=0.45\textwidth]{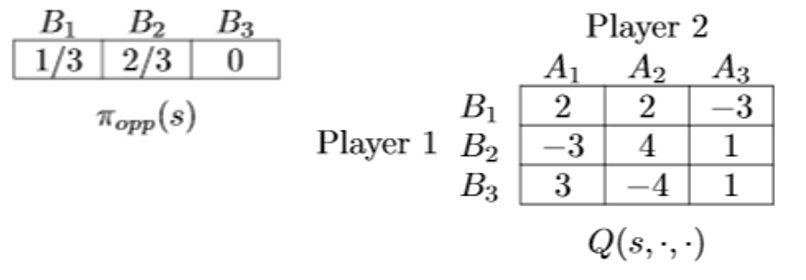}
    \caption{Policy times $Q(s, \cdot, \cdot)$ value matrix in 3-action game.}
    \label{fig:policy_times_q}
\end{figure}

Pseudocode for BRAT is given in Algorithm 2. The first difference between the Minimax-DQN algorithm and BRAT is that we train against a particular opponent instead of in self-play. We thus construct a classification model of this particular opponent's policy, $\pi^{opp}_{\phi}$, and periodically update it using log-loss. BRAT uses the model $\pi^{opp}_{\phi}$ to exploit our opponent by choosing actions greedily as
\begin{equation}
    a_t = \underset{a_t}{\text{argmax}} \hspace{.2cm} \pi^{opp}_{\phi}(s_t) Q_\theta(s_t, a_t, \cdot).
\end{equation}
Here the policy $\pi^{opp}_{\phi}(s_t)$ is a vector of probabilities for each action given by our classification model, and we multiply by whichever column of the Q value matrix $Q_\theta(s_t, \cdot, \cdot)$ that will yield the highest expected value. Figure \ref{fig:policy_times_q} shows an example of  this vector-matrix product for a 3-action game.  

\noindent In Algorithm 2, the last notable difference is that the TD-target is set as either
\begin{align*}
	y_i &= r_{t,i} + \gamma \cdot \underset{a}{max}  \left[ \pi^{opp}_{\phi}(s_{t+1,i}) Q_\theta^-(s_{t+1, i}, a,\cdot)\right] \\
	& \hspace{3cm} OR \\
    y_i &= r_{t,i} + \gamma \cdot \underset{a}{\max} \hspace{.1cm}\underset{b}{\min}  \left[ Q_\theta^-(s_{t+1, i}, a,b)\right].
\end{align*}
\noindent The first option is more greedy; we assume the value of being in subsequent state $s_{t+1}$ is dependent on our model of our opponent's policy, and act accordingly. The second option assumes that our opponent will act optimally from the subsequent state. We found empirically that the second option is generally more effective, although computing the min-max value of the game can be much more computationally expensive, especially for large action spaces.

\section{Meta-Nash DQN}
The second agent we introduce builds an \textit{implicit} model of the opponent using a GRU to create a context variable $C$ which is in turn used to compute matrix values $Q(C, s, \cdot, \cdot)$. To select actions, this agent directly computes a Nash-Equilibrium policy for the particular output $Q(s, \cdot, \cdot)$.

The implicit model of the opponent's policy is constructed using (state, action) pairs, similar to BRAT. During execution, the pairs $(S_{t-1}, a_{t-1})$ are continually fed into the GRU to produce a context variable $C_{t-1}$. This context variable is fed into the DQN, along with the current state, to produce matrix values $Q(C_{t-1}, s, \cdot, \cdot)$. The Nash Equilibrium set of policies are explicitly calculated with the Lemke-Howson algorithm, and an action is sampled from the mixed Nash Equilibrium policy for the Meta-Nash Agent. The contiguous trajectories are stored in the replay buffer along with the context variables $C$ and hidden states $h$ which are needed to learn the parameters of the GRU. 

During training, random chunks of the trajectories are sampled of length equal to a \textit{GRU-length} hyperparameter, and fed into the GRU along with the correct hidden state obtained from the replay buffer to produce context $C_{t-1}$. TD-targets are computed using the min-max Nash Equilibrium values similar to Minimax-DQN, but now also using the context variable from the replay buffer. The TD-error is both a function of the parameters $\phi$ of the GRU and the parameters $\theta$ of the DQN: 
\begin{align*}
    L(&\theta, \phi) =
    \mathrm{E}_{(s,a,b,s',r) \sim D} \biggl[
    \\
    & \left( Q_\theta(C^{\phi}, s, a, b) - r + \gamma \underset{a'}{\max}\underset{b'}{\min}Q_\theta(C^{\phi},s',a', b') \right)^2\biggr].
\end{align*}
The full pseudocode for the Meta-Nash Agent is given in Algorithm \ref{alg:Meta-Nash}.

\begin{algorithm}[h!]
\begin{algorithmic}
\State \textbf{Input}: A two player zero-sum Markov game ($\mathcal{S}, \mathcal{A}, \mathcal{B}, P, R, \gamma$), replay buffer $\mathcal{D}$, minibatch size $n$, set of opponent policies $\pi_i$, and architecture of minimax deep Q-network $Q_\theta: \mathcal{S} \times \mathcal{A} \times \mathcal{B} \xrightarrow{}{}\mathcal{R}. $
\State Initialize replay memory $\mathcal{D}$ to capacity $N$
\State Initialize action-value network $Q_\theta$ with random weights $\theta$
\State Initialize target network $Q_\theta^*$ = $Q_\theta$ 
\\
\State Initialize context GRU with parameters $\phi$ and set \textit{GRU-Length}

\While{not done}
\State Sample batch of policies $\pi_i$
    \For{each $\pi_i$}
        \For {$k=1,K$}
        	\State Select opponent action $a_t \sim \pi_i$ 
        	\State Select own action 
        	\State \hspace{1cm} $\displaystyle b_t \sim \max\min Q_\theta(C_{t-1}, S_t, a_t, b_t)$
        	\State Execute actions $a_t$, $b_t$ and observe $R_t$, $S_{t+1}$
        	\State Feed transition through GRU
        	\State \hspace{1cm}$ \displaystyle h_t, C_t = GRU_{\phi} (h_{t-1}, (S_t, a_t))$
        	\State Append transition \State $\left(h_{t-1}, C_t, S_t, a_t, b_t, R_t, S_{t+1}\right)$ 
        	\State To replay buffer $\mathcal{D}_i$
        	\EndFor
        	\State Sample minibatch of transitions from $\mathcal{D}_i$
        	\State Compute adapted paramters by setting the target
        	\State \hspace{.1cm} $y_i = R_t + \gamma \cdot \underset{a}{\max}\underset{b}{\min}  \left[Q_\theta^*(C_{t,i}, S_{t+1, i}, a, b)\right]$
        	\State and then updating parameters
             \State \hspace{.1cm}	$\theta, \phi \leftarrow \argmin \sum_{i \in [n]}\left(y_i - Q_\theta(C_{t-1, i}, S_{t,i},a_{t,i}, b_{t,i})\right)^2$
    \EndFor
\EndWhile
\end{algorithmic}
\caption{Meta Learning Nash-DQN}
\label{alg:Meta-Nash}
\end{algorithm} 

This algorithm is called \textit{\textbf{Meta}}-Nash DQN because there is a nice parallel between opponent modelling using a GRU and the work of Meta-Q-Learning~\cite{fakoor_meta-q-learning_2020}. If we consider each opponent we train against as a distinct task, then Fakoor et. al.'s Meta-Q-Learning would feed in transitions $(s, a, s', r)$ into a GRU to produce a context variable which in turn is used to produce Q-values. Unlike Meta-Q-Learning, we only need the context variable to contain information about our opponent's policy, and not the reward or transition functions and thus we only feed in $(s, a)$ pairs into the GRU. Furthermore, we want to be able to model our opposing agent as a nonstationary entity instead of part of the environment, and therefore handle a matrix of Q-values.

\section{Related Work}
Literature on multi-agent reinforcement learning (MARL) and competitive games has recently surged in popularity. Early attempts approached multi-agent problems with single agent learners that treat other agents as part of the environment. However, this has been found not to work well in practice, likely due to the nonstationarity of each agent's environment~\cite{matignon2012independent}.  There has been extensive work focusing explicitly on two player games in strategic turn based setting such as Go, Chess, and Backgammon~\cite{tesauro1994td, silver2018general}. There has been much less focus on the two player simultaneous action games that we address. 

An important algorithm in the history of MARL developed by Lowe et. al. is Multi-Agent Deep Deterministic Policy Gradients (MADDPG)~\cite{lowe2017multi}. MADDPG uses a centralized Q-value $Q_i^{\boldsymbol\pi}(s, a_1, ..., a_n)$ function for each agent $i$ which takes as input the state and actions of each agent, and outputs the Q-value for agent $i$. Lowe et al. derive an actor critic method similar to DDPG to learn a policy from this value function in continuous action spaces. However, to update the Q-value function requires the policy of other agents. Lowe et al. suggest using maximum likelihood with entropy regularization to model other agents' policies when they are unknown.

Recently Li. et al. proposed an extension to MADDPG called Minimax Multi-Agent Deep Deterministic Policy Gradients (M3DDPG)~\cite{li_robust_2019}. This algorithm adds the assumption that all other agents are acting adversarially, and thus takes a minimum over all opponent actions in Q-value updates. However, they note that finding the min over all other agents actions is computationally intractable, especially as they work with continuous action spaces. Thus they use a linear approximation of the Q-function and take one gradient step towards the action that minimizes the Q-value. This algorithm is a nice extension of the Minimax-DQN algorithm from~\cite{fan_theoretical_2020} to continuous action spaces with multiple agents. However, like the Minimax-DQN algorithm, M3DDPG forces the agent to act conservatively with no way to exploit a suboptimal opponent.

There have also been many efforts in MARL that use a central value network (either $V(s)$ or $Q(s, a)$) which can be decomposed into individual value functions for each agent~\cite{sunehag2018value, rashid_qmix_2018}. However, these methods have only been used in cooperative settings. One of the main motivations of such an approach in multi-agent scenarios is that again the joint action space grows exponentially in the number of agents, and thus planning with a classic central value network is intractable for many problems. These methods have proved effective in cooperative games such as team searching and fetching and Starcraft unit micromanagement~\cite{foerster_counterfactual_2017}.

Michael Bowling and Manuela Veloso introduced two algorithms for handling two player simultaneous action games using tabular Q-value functions and policies~\cite{bowling2001rational}. The first simple algorithm called policy hill climbing (PHC) is sort of tabular actor critic method that uses mixed strategies. The second algorithm called Win or Lose Fast PHC (WoLF PHC) adds a variable learning rate to the PHC algorithm which is large while WoLF PHC is losing and small when WoLF PHC is winning, where winning/losing is evaluated by comparing the most recent returns to average returns.

A more recent paper that handles two player simultaneous action games is Learning with Opponent Learning Awareness (LOLA)~\cite{foerster_learning_2018}. LOLA is a policy gradient algorithm that uses the parameters of their opponent to make policy updates based on a forecast of their opponent's learning. To do so, Foerster et. al assume that they have access to their opponents policy parameters, which the authors admit is a unrealistic assumption. They develop a weaker version of LOLA which does not make this assumption, and instead uses a maximum likelihood estimate to infer their opponents policy, similar to MADDPG. 

\section{Experiment Methodology}
\label{sec:experiments}
\begin{figure}[h!]
    \centering
    \includegraphics[width=0.45\textwidth]{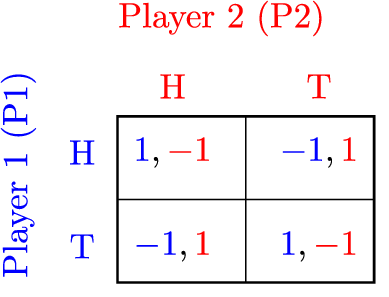}
    \caption{Payoff Matrix for Matching Pennies Game}
    \label{fig:matching-pennies-payoffs}
\end{figure}

\begin{figure}[h!]
    \centering
    \includegraphics[width=0.45\textwidth]{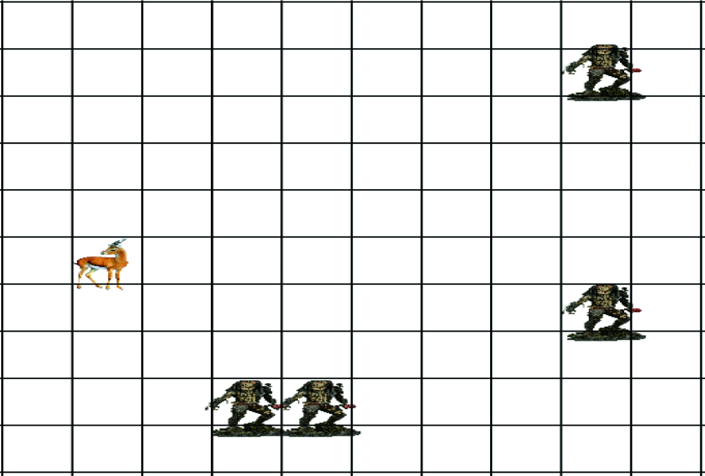}
    \caption{Sample environment for predator-prey game}
    \label{fig:pred-prey-env}
\end{figure}

We consider tests in two different environments: (1) a simple iterated matrix game of Matching Pennies, and (2) a predator-prey game. The first environment consists of repeatedly playing the matrix game shown in Figure \ref{fig:matching-pennies-payoffs}. The second environment takes place in a gridworld as depicted in Figure \ref{fig:pred-prey-env}, where some agents are predators, and other are prey. The predators get +1 reward whenever they catch a prey, and the prey get -1 reward whenever they are caught. The discount factor $\gamma$ is important in this setting so that predators are incentivized to quickly catch the prey, and the prey are incentivized to survive for as long as possible.

In both environments we set up a tournament that consists of multiple agents competing in a round-robin setting. We consider six agents; the Best Response AgenT (BRAT), the Meta-Learning NashDQN (Meta-Nash), Win or Lose Fast Policy Hill Climbing (WoLF-PHC), single agent Q-Learning (Q-Learning), single agent Policy Gradient (PG), and Learning with Opponent Learning Awareness (LOLA). During the round robin tournament, all pairs of agents play some number of matches/episodes against eachother. For the Iterated Matching Pennies environment, the matches/episodes consist of 100 transitions. A total of 50 episodes are run consecutively. This is done for 25 different random trials (different parameter initialization).

For the predator-prey game we use a similar tournament setup, however the number of transitions per episode generally does not reach the maximum of 100 transitions, as the prey is caught sooner. Since the episodes are typically shorter, we run for a total of 200 different episodes. We experiment with both a 1v1 predator-prey game and a 3v1 predator-prey game where we take the joint action space of the 3 predators to reduce the problem to a two agent MDP.

Before the round robin tournaments there is the option to pretrain the agents. Notice that the algorithms we develop, and in particular the Meta-Nash DQN are designed to train against many different agents in order to be able to quickly adapt to playing a new agent (the simple behavioral cloning of BRAT does not actually use/require this setup, but ideally a few-shot imitation learning algorithm that did utilize this pretraining would be used instead of behavioral cloning). Other algorithms typically do not have a pretraining step like this. For the Iterated Matching Pennies tournament we do not pretrain, but for the predator-prey environment we do pretrain each agent by allowing all pairs of agents to play for 50 episodes before results are started to be recorded (only a single set of parameters for each agent are used during this step, while during the tournament each agent has a different set of parameters for each match).

We restrict our testing to smaller environments where the success of each agent's action is meaningfully dependent on the opponent's action.
We found empirically that the algorithms we developed did not scale well to games with larger state and action spaces such as the Google Research Football environment~\cite{kurach_google_2020}.

\section{Results}

\begin{figure}[h!]
    \centering
    \includegraphics[width=0.51\textwidth]{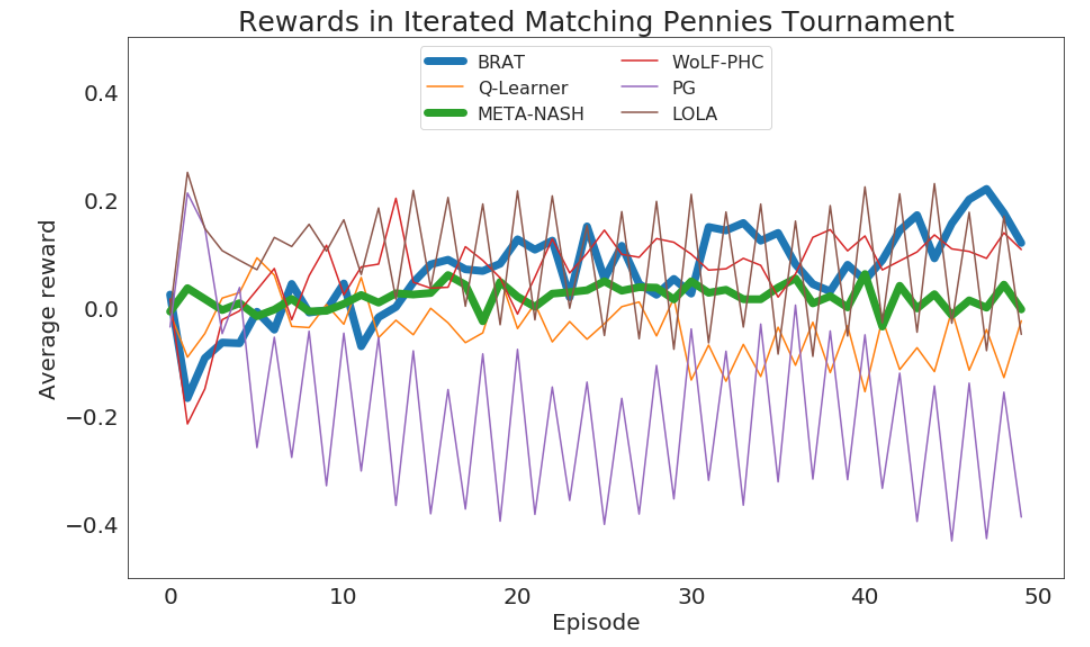}
    \caption{Graph for rewards in IMP}
    \label{fig:IMP-line}
\end{figure}

The average rewards for the Iterated Matching Pennies tournament are plotted in Figure \ref{fig:IMP-line}. The reward values are averaged over the five games each agent plays, one game against each opposing agent, as well as averaged over 25 different parameter initializations. Notice that the Meta-Nash algorithm consistently receives rewards near 0 for the average reward in each episode. This is likely due to the explicit minimax calculation over Q-values which forces the agent to play closely to Nash Equilibrium. On the other hand, the BRAT agent slowly improves performance, which we speculate is due to the fact that the BRAT agent is gradually building a more accurate model of its opponent.

\begin{figure}[h!]
    \centering
    \includegraphics[width=0.51\textwidth]{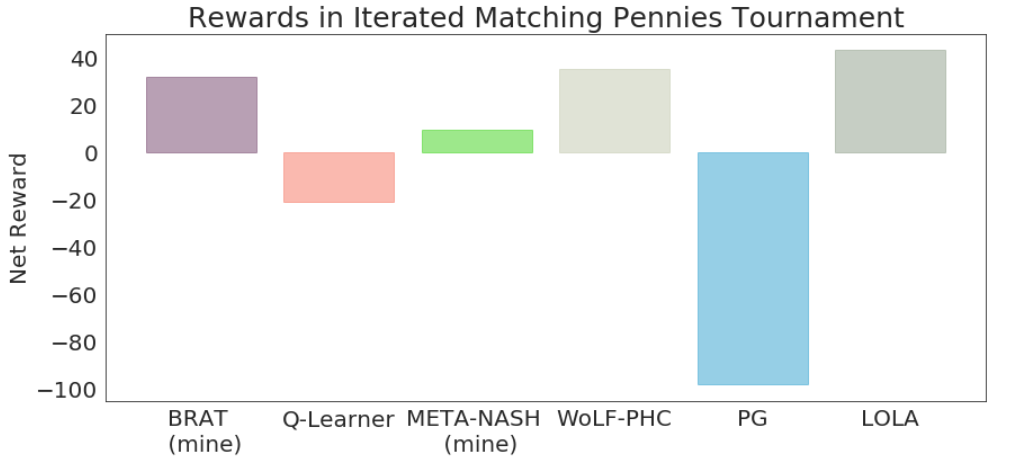}
    \caption{Bar graph for cumulative rewards in IMP}
    \label{fig:IMP-bar}
\end{figure}

In Figure \ref{fig:IMP-bar}, we again plot the average rewards for the IMP tournament, this time as a bar graph of the cumulative reward over all 50 episodes. The BRAT and Meta-Nash agents perform competitively with other state of the art methods.

\begin{figure}[h!]
    \centering
    \includegraphics[width=0.49\textwidth]{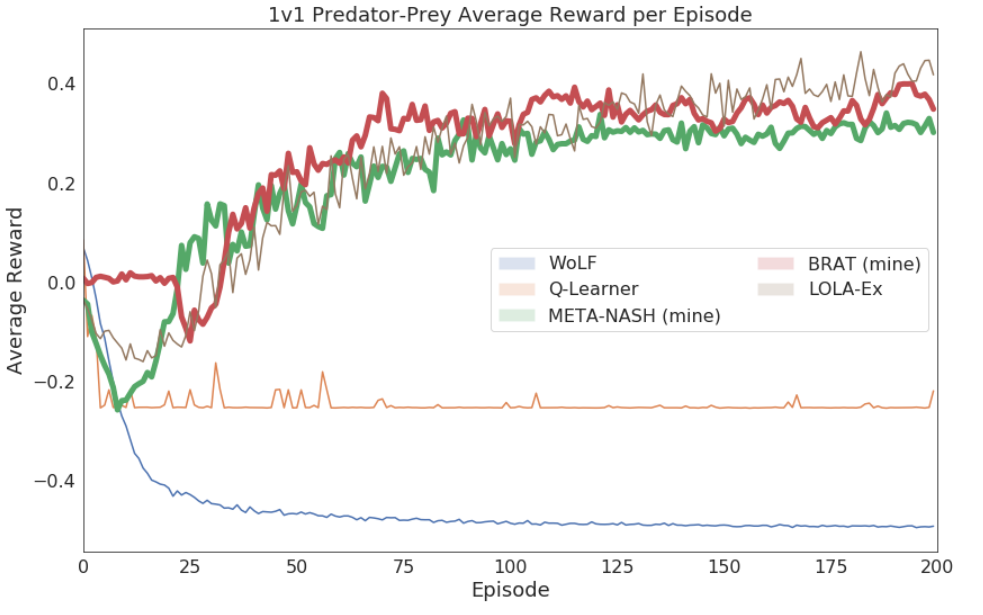}
    \caption{Results from 1v1 predator-prey tournament}
    \label{fig:1v1-pred-prey}
\end{figure}

The average rewards for the 1v1 predator-prey tournament are plotted in Figure \ref{fig:1v1-pred-prey}. Again we are plotting the average reward per episode which is averaged over the five different games the agent is playing and 25 different random initializations.
\newline
\begin{figure}[h!]
    \centering
    \includegraphics[width=0.49\textwidth]{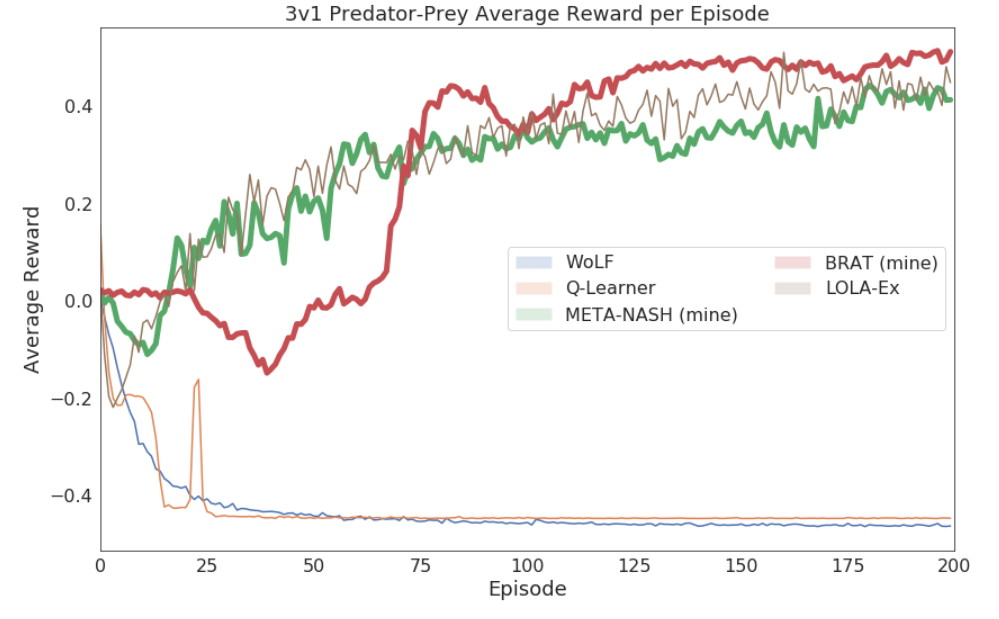}
    \caption{Results from 3v1 predator-prey tournament}
    \label{fig:3v1-pred-prey}
\end{figure}


\bibliography{Multi-Agent-RL.bib}

\section{Appendix A: Code Breakdown}
Code for this paper is available at 
\url{https://github.com/peweetheman/Reinforcement_Learning_In_Two_Player_Simultaneous_Action_Games}.
To run any of the code first use a package manager to install the packages from \textit{requirements.txt}. The other files exposed in the main folder include the code for the Meta-Nash agent and the BRAT agent, as well as the code to run the tournaments. The \textit{run\_tournament} file is currently set up to run the Iterated Matching Pennies (IMP) tournament for 25 randomly initialized trials each consisting of 50 episodes, each episode consisting of 100 transitions. The IMP tournament was run with $\gamma=1.0$ and using Adam optimizer with $lr=0.005$. These were the same settings used to generate the results in this paper.

To run tournaments for the predator-prey environment, there is a conflicting dependency for the \textit{gym} package. I believe the only change that has to be made is to replace gym version .17.0 with .10.0, and then the predator-prey environment from the folder \textit{ma-gym} should be available. 

To plot the results from the tournaments, look in the folders \textit{matching-pennies\_tournament} and \textit{predator-prey\_tournament}. In each folder there is data saved in a folder with the environment name, a .ipynb with the code to load the data and plot, and pdf images of the results.

The file \textit{CNN\_context\_dqn\_nash\_eq} contains a version of the Meta-Nash algorithm suitable for environments that have a pixel state space such as the atari environments. No tests of this environment were included in this report.

The $\textit{common}$ folder includes some functionality that is shared in various algorithms implemented such as the replay buffer code.
\end{document}